\documentclass[conference]{IEEEtran}
\IEEEoverridecommandlockouts
\usepackage{cite}
\usepackage{amsmath,amssymb,amsfonts}
\usepackage{algorithmic}
\usepackage{graphicx}
\usepackage{textcomp}
\usepackage{xcolor}

\usepackage{multirow}
\usepackage{multicol}
\usepackage{url}
\def\BibTeX{{\rm B\kern-.05em{\sc i\kern-.025em b}\kern-.08em
    T\kern-.1667em\lower.7ex\hbox{E}\kern-.125emX}}
\begin{document}

\title{You Only Landmark Once: Lightweight U-Net Face Super Resolution with YOLO-World Landmark Heatmaps}


\IEEEoverridecommandlockouts 

\author{
\IEEEauthorblockN{
    Riccardo Carraro\textsuperscript{*}\thanks{\textsuperscript{*}Indicates equal contribution}, 
    Anna Briotto\textsuperscript{*}, 
    Endi Hysa, \\
    Marco Fiorucci, and 
    Lamberto Ballan
}
\IEEEauthorblockA{
    \textit{University of Padua}, Padua, Italy \\
    \{riccardo.carraro.10, anna.briotto, endi.hysa\}@studenti.unipd.it, \\
    \{marco.fiorucci, lamberto.ballan\}@unipd.it}
}

\maketitle

\maketitle

\begin{abstract}
Face image super-resolution aims to recover high-resolution facial images from severely degraded inputs. Under extreme upscaling factors, fine facial details are often lost, making accurate reconstruction challenging. Existing methods typically rely on heavy network architectures, adversarial training schemes, or separate alignment networks, increasing model complexity and computational cost.
To address these issues, we propose a lightweight U-Net based-architecture designed to reconstructs $128{\times}128$ facial images from severely degraded $16{\times}16$ inputs, achieving an $8\times$ magnification. 
A key contribution is a novel auxiliary-training-free supervision strategy that leverages heatmaps generated by YOLO-World, an open-vocabulary object detector, to localize key facial features such as eyes, nose, and mouth. 
These heatmaps are converted into spatial weights to form a heatmap-guided loss that emphasizes reconstruction errors in semantically important regions. Unlike prior methods that require dedicated landmark or alignment networks, our approach directly reuses detector outputs as supervision, maintaining an efficient training and inference pipeline. Experiments on the aligned CelebA dataset demonstrate that the proposed loss consistently improves quantitative metrics and produces sharper, more realistic reconstructions. Overall, our results show that lightweight networks can effectively exploit detection-driven priors for perceptually convincing extreme upscaling, without adversarial training or increased computational cost.

\end{abstract}

\begin{IEEEkeywords}
Face Super-resolution, Lightweight image super-resolution, Open-vocabulary priors
\end{IEEEkeywords}

\section{Introduction}
Single Image Super-Resolution (SISR) is a fundamental task in Computer Vision, aiming to reconstruct a high-resolution image from its low-resolution counterpart. High-quality super-resolution is widely applicable, including media restoration, enhancement of visual content, and face recognition in surveillance and digital forensics~\cite{park2003super,yang2019deep}. In these scenarios, the recovery of fine details in critical regions strongly influences the perceptual quality of the reconstructed image.
In the context of face image super-resolution, accurately reconstructing local facial structures, such as eyes, mouth, and nose, is crucial to maintaining both identity and visual realism. To this end, recent approaches have moved beyond traditional generative models \cite{ledig2017photo,lugmayr2020srflow,wang2018esrgan}, towards computationally heavy diffusion models \cite{wu2024diffbir,zhang2024seesr}, large-scale vision transformers \cite{chen2023hat,zhang2024atd}, attention mechanisms \cite{shi2025face}, or specialized landmark-based pipelines to guide the reconstruction of these key regions. While these methods often achieve impressive visual quality, they typically come at the cost of prohibitive model complexity, extensive training requirements, and limited applicability in resource-constrained or real-time settings. Several methods have explored the integration of landmark or attention information to guide the restoration of key regions of faces. However, deriving such guidance often requires dedicated face alignment networks or heatmap generators \cite{bulat2018super,kim2019progressive,chen2018fsrnet}, which can introduce complexity and pose challenges for generalization and adaptation across different application domains. Some of these approaches, in particular \cite{bulat2018super}, exhibit limited robustness to occluded landmarks, resulting in degraded reconstruction quality and visual artifacts.

In this work, we propose an approach that mitigates these critical points by using heatmaps extracted from YOLO-World zero shot detections~\cite{yoloworld2024}. This choice eliminates the need for training an auxiliary network and ensures that attention is focused only on visible features detected in the image.
From an architectural perspective, we deliberately adopt a minimal and lightweight design, allowing us to focus on the effect of the proposed priors rather than architectural complexity. 


Instead of introducing the complexity of previous methods ~\cite{ledig2017photo, zhang2024seesr, ma2020deep,shi2025face,wang2023sfmnet}, we adopt a U-Net backbone~\cite{ronneberger2015u}. This choice provides improved training stability, faster convergence, and a favorable balance between reconstruction quality and computational efficiency, making the approach well suited for efficiency-critical scenarios.
We evaluate our method on the aligned CelebA dataset~\cite{liu2015faceattributes} under the same bilinear degradation protocol used by \cite{chen2018fsrnet,kim2019progressive,kim2016accurate,lu2022edge}, enabling direct comparison with prior landmark-guided approaches. In addition, to ensure comparability with more recent face super-resolution methods, we also report results under bicubic downsampling settings \cite{ma2020deep,shi2025face,wang2023sfmnet}.


In summary, our contributions are: i) Introducing a novel method to obtain priors for face super-resolution, obtained at zero additional training cost from the general-purpose YOLO-World model, which improves the reconstruction of key facial regions without relying on landmark detectors or auxiliary supervision; ii) presenting a lightweight and stable super-resolution framework based on a U-Net backbone, offering a favorable balance between reconstruction quality and computational efficiency; and iii) validating the proposed approach through extensive experiments on the aligned CelebA dataset under both bilinear and bicubic degradation settings, demonstrating its robustness and general applicability.


\section{Method}
\subsection{Architecture}
\begin{figure*}
    \centering
    \includegraphics[width=1\linewidth]{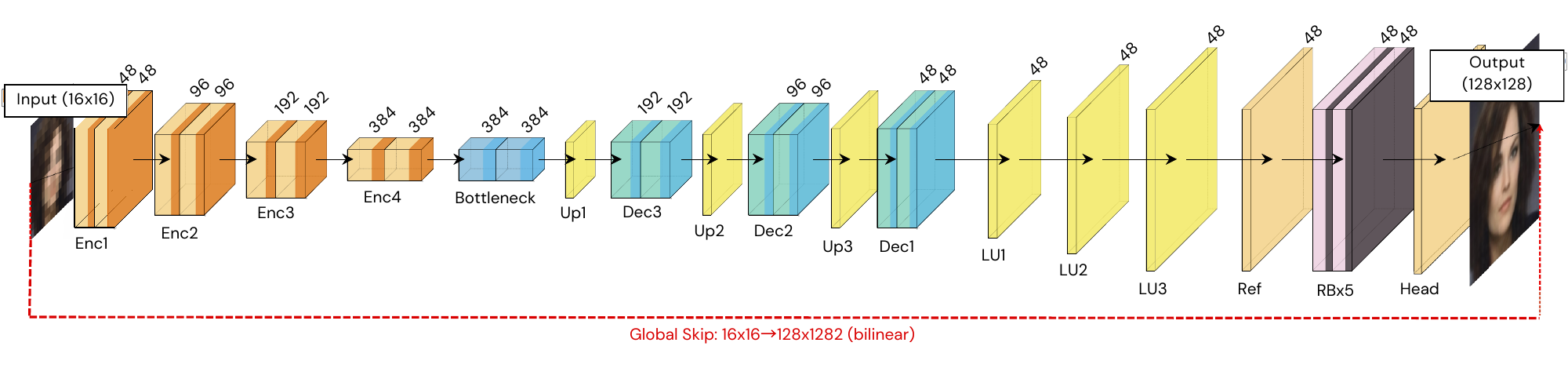}
    \caption{Efficient U-Net architecture for image super-resolution, transforming a low-resolution $16\times16$ input into a high-resolution $128\times128$ output with $8\times$ upsampling. The encoder blocks (Enc) halve the spatial size at each stage, while the decoder blocks (Dec) and upsampling blocks (Up, LU) progressively restore and increase resolution. The refinement stage (Ref, RBx5) further enhances spatial details. A global skip connection (bilinear/bicubic interpolation) adds the upsampled input to the output for stability.}
    \label{fig:architecture}
\end{figure*}

We propose a lightweight U-Net for face super-resolution, mapping $16{\times}16$ inputs to $128{\times}128$ outputs (Figure~\ref{fig:architecture}). The model follows an encoder-bottleneck-decoder design~\cite{ronneberger2015u}, with modifications aimed at improving detail preservation and output fidelity. 

The encoder is composed of four convolutional blocks with increasing channel depth, starting from 48 filters. Each block applies two $3{\times}3$ convolutions with LeakyReLU activations ($\alpha{=}0.2$), using stride-2 convolutions for downsampling instead of max-pooling~\cite{radford2015unsupervised}. The bottleneck further processes the compressed representation with two additional convolutional layers. The decoder mirrors the encoder through progressive upsampling (bilinear or bicubic, matching the input degradation method), skip connections, and double-convolution blocks~\cite{ronneberger2015u}. After recovering the original spatial resolution ($16{\times}16$), a learned upsampling module progressively increases it through three stages, $16{\to}32{\to}64{\to}128$. Each stage applies nearest-neighbor interpolation followed by a $3{\times}3$ convolution and LeakyReLU, avoiding the checkerboard artifacts often associated with transposed convolutions~\cite{dong2016accelerating}. At full resolution ($128{\times}128$), a refinement head with five residual blocks~\cite{he2016deep} further improves the reconstruction. A $1{\times}1$ convolution reduces the feature dimensionality before the residual stack, and a final $1{\times}1$ layer projects the features to RGB output. A global skip connection adds an upsampled version of the low-resolution input to the final prediction, improving convergence and color consistency~\cite{lim2017edsr}. The full model contains approximately 7.3 million parameters.

\subsection{Heatmap-based loss}
\label{heatmap loss}
Our approach involves computing heatmaps for the entire dataset and using them as weighted masks over the prediction errors, thereby assigning greater penalties to errors in critical face landmark. This encourages the model to focus on accurately predicting the most important regions of the face.
To this end, we first use YOLO-World~\cite{yoloworld2024} to detect facial components in the target images and obtain coarse, semantically meaningful regions of interest, which are subsequently used to generate pixel-aligned importance heatmaps. Unlike ~\cite{kim2019progressive}, which rely on distilled FAN landmarks, our method leverages YOLO-World’s open-vocabulary framework to define prompts for facial landmark detection, removing the need for a task-specific auxiliary network.
Building upon these detected regions, we construct fine-grained, structure-aware heatmaps that capture the spatial extent and the key geometry of each facial component. Specifically, each detected region is cropped and processed to enhance structural details: we apply Scharr and Canny edge detectors, normalize the responses, and smooth them with a Gaussian blur (kernel size $15 \times 15$, $\sigma=3$). To avoid assigning uniform importance across the entire bounding box, we modulate the edge maps with spatial fading functions with respect to the pixel coordinates \((x,y)\), normalized to \([-1,1]\) relative to the box center:
\begin{itemize}
    \item Gaussian fade: $f(x,y) = \exp(-2(x^2+y^2))$, emphasizing the center of small features (e.g., eyes, nose, mouth).
    \item Inverse Gaussian fade: $f(x,y) = 1 - \exp\!\left(-\tfrac{x^2+y^2}{2\sigma^2}\right)$ with $\sigma=0.6$, emphasizing the contours of larger regions (e.g., face, head).
\end{itemize}

\begin{figure}
    \centering
    \includegraphics[width=0.55\linewidth]{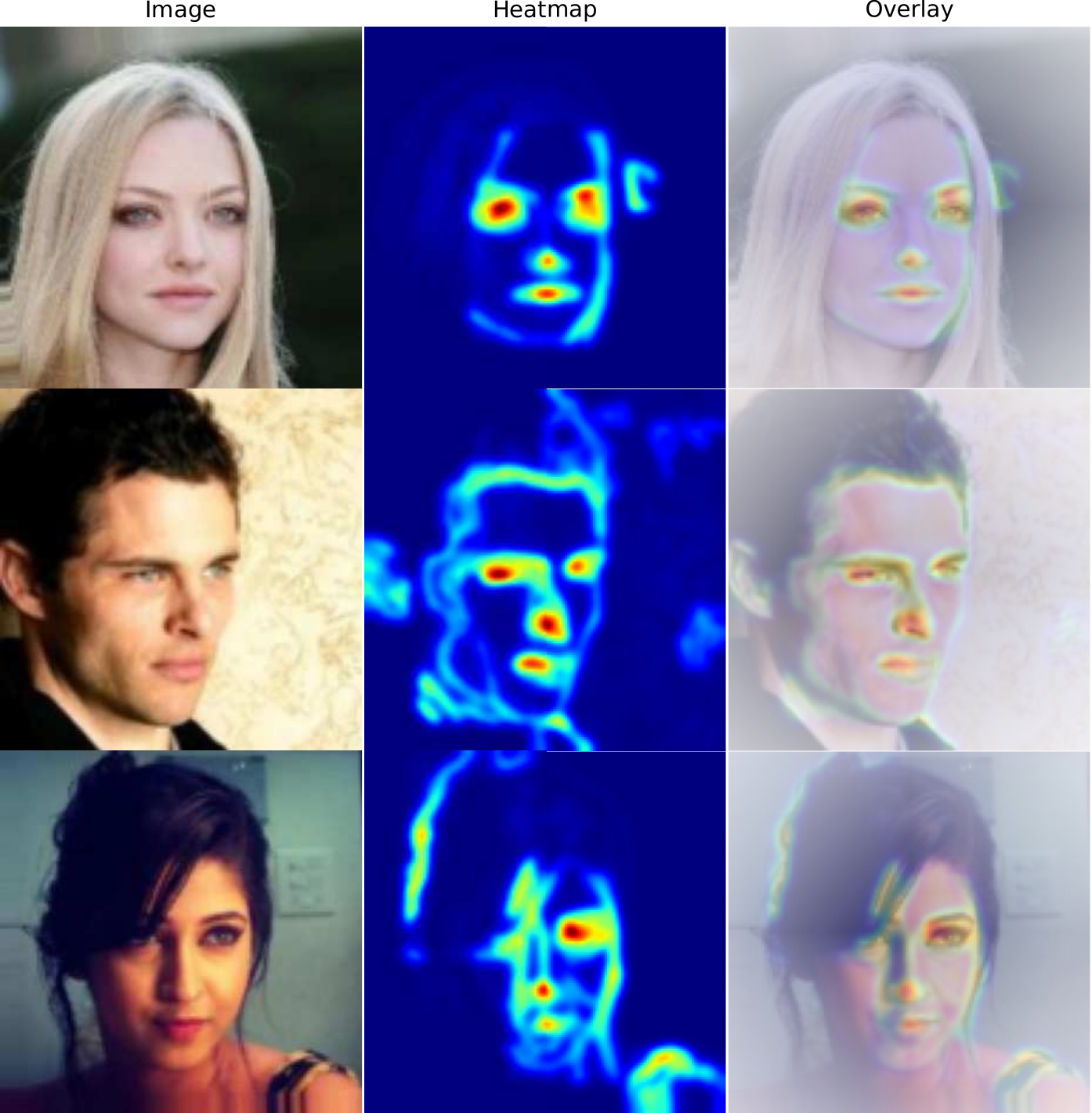}
    \caption{Heatmaps generatad with landmarks detected by YOLO-World.}
    \label{fig:heatmaps}
\end{figure}

Class-specific weights were manually assigned based on qualitative inspection of the output heatmaps, with the goal of emphasizing visually prominent facial regions. Specifically, we assign a weight of 4.5 to the eyes; 4.0 to the eyebrows, mouth, nose, chin, ears, and face; 3.0 to the nose tip; and 2.0 to the head.
The contributions from all detections are accumulated into a combined and normalized heatmap that matches the resolution of the target image, as shown in figure~\ref{fig:heatmaps}. This formulation enables the direct use of the heatmap as weighting masks in the reconstruction loss. Specifically, we define a weighted pixel loss that reweights reconstruction errors according to the pre-computed YOLO-World heatmaps. Let $H \in [0,1]^{H \times W}$ be the normalized heatmap. The heatmap weighted loss is:

\begin{equation}
\mathcal{L}_{\text{heat}} = 
\frac{1}{N} \sum_{i=1}^{N} 
\frac{w_i \cdot \lvert \hat{I}_i - I_i \rvert}{\sum_{j=1}^{N} w_j},
\end{equation}

where $w_i = \textit{floor} + (1-\textit{floor}) \cdot H_i^\gamma$, 
$\hat{I}_i$ and $I_i$ are the predicted and ground-truth pixel intensities, $N$ is the total number of pixels, $\gamma > 1$ controls the selectivity of hot regions, and \textit{floor} $\in (0,1]$ ensures non-zero weight outside the salient areas. 

In this way, errors in high-importance regions (e.g., eyes, mouth) contribute more strongly to the total loss, while errors in less relevant areas are still considered but down-weighted.

\subsection{Composite Loss Function}
To train our super-resolution network, we employ a composite loss function that balances pixel accuracy, perceptual similarity, semantic awareness, and visual realism. The overall loss is defined as:
\begin{equation}
\mathcal{L}_{\text{total}} = \mathcal{L}_{\text{pix}} + \lambda_{\text{perc}} \mathcal{L}_{\text{perc}} + \lambda_{\text{heat}} \mathcal{L}_{\text{heat}} + \lambda_{\text{lpips}} \mathcal{L}_{\text{lpips}},
\end{equation}
where each component is scaled by a tunable weight $\lambda$.

\begin{itemize}
    \item \textbf{Pixel Loss:} We use the standard mean squared error (MSE) to enforce low-level fidelity between the predicted and ground-truth images:
    $\mathcal{L}_{\text{pix}} = \frac{1}{N} \sum_{i=1}^{N} \lVert \hat{I}_i - I_i \rVert^2$,
    where $\hat{I}_i$ and $I_i$ denote the predicted and ground-truth pixel intensities, and $N$ is the number of pixels.
    
    \item \textbf{Perceptual Loss:} Following ~\cite{kim2019progressive}, to capture high-level semantic structure, we adopt a perceptual loss computed on feature activations of a pretrained VGG16 network, following~\cite{johnson2016perceptual}. Specifically, we compare features from conv1\_2, conv2\_2, and conv3\_3 using MSE:
    $\mathcal{L}_{\text{perc}} = \sum_{l=1}^{3} \lVert \phi_l(\hat{I}) - \phi_l(I) \rVert^2$,
    where $\phi_l(\cdot)$ denotes the feature extractor at layer $l$. Inputs are normalized to ImageNet statistics after mapping from $[-1,1]$ to $[0,1]$.
    
    \item \textbf{Heatmap Loss:} $\mathcal{L}_{\text{heat}}$ is defined in section \ref{heatmap loss}.
    
    \item \textbf{LPIPS Loss:} We also incorporate the Learned Perceptual Image Patch Similarity (LPIPS)~\cite{zhang2018lpips}, which compares deep features extracted from VGG networks, fine-tuned to match human perceptual judgments. By encouraging similarity in deep feature space, LPIPS enhances the perceptual realism of reconstructed images, promoting sharper textures and more human-aligned reconstructions. This aligns with the growing trend in super-resolution literature that shifts focus from signal fidelity to perceptual quality~\cite{wang2018esrgan}.
\end{itemize}

\section{Experiments}
\subsection{Dataset}
We conduct our experiment on the aligned version of the CelebA dataset~\cite{liu2015faceattributes}, a large-scale collection of 202.599 celebrity face images with variability in pose, background and illumination. Following \cite{chen2018fsrnet,kim2019progressive,kim2016accurate,lu2022edge,lugmayr2020srflow}, each $128{\times}128$ HR image is bilinearly downsampled to $16{\times}16$ to generate the LR input, ensuring comparability with prior work. To align our experimental protocol with contemporary face super-resolution methods~\cite{ma2020deep,shi2025face,wang2023sfmnet}, we also train and evaluate the proposed pipeline under bicubic downsampling of the HR images.
We used the full dataset in our experiments, comprising of 162.770 images for training, 19.867 for validation and 19.962 for testing.

\subsection{Metrics}
\noindent\textbf{Restoration quality evaluation.} Following standard face super resolution works, we evaluate the quality of the reconstructed images using Peak signal-to-noise ratio (PSNR), Structural Similarity Index (SSIM)~\cite{wang2003multiscale}, and Multi-scale Structural Similarity (MS-SSIM).
PSNR is a pixel-wise fidelity measure, while SSIM accounts for local structural consistency by considering luminance, contrast and texture. MS-SSIM~\cite{wang2003multiscale} extends SSIM to multiple resolutions, providing a more robust assessment of visual quality in line with human perception. This combination of metrics ensures that both low-level fidelity and perceptual realism are properly captured, allowing direct comparison with prior work. 
In addition, we perform qualitative inspection on validation and test outputs to assess perceptual realism and semantic consistency.

\noindent\textbf{Computational efficiency analysis.}
We also evaluate the computational efficiency of our model by reporting Multiply-Accumulate operations (MACs), a hardware-agnostic measure of the number of operations required for a single forward pass. Following recent work on efficient image super-resolution\cite{hui2019imdn,liu2020rfdn,zhang2023bsrn}, MACs are computed at inference time with a fixed low-resolution input of $16 \times 16$ pixels and an upscaling factor of $8\times$.
In addition to MAC-based complexity, we report the number of trainable parameters, inference latency (ms) and throughput (FPS). Runtime measurements are obtained by averaging over 200 forward passes on a fixed input resolution of $16\times16$ pixels, with an $8\times$ upscaling factor, ensuring a fair and reproducible evaluation of practical efficiency.

\subsection{Training Details}
Our model is trained and tested on the aligned CelebA dataset using low-resolution inputs of $16{\times}16$ and high-resolution ground truths of $128{\times}128$. We use a batch size of 8 and train using early stopping on a single Nvidia RTX A5000 GPU with mixed-precision training enabled. Reproducibility is ensured by fixing seeds and enforcing deterministic behavior\cite{henderson2018deep}. Training one single epoch on the whole dataset (training and validation together) required about 20 minutes. All the models reached convergence within 43 epochs. In addition to metric logging, validation samples were visually inspected every 5 epochs to ensure that training improvements translated into perceptually better reconstructions. The complete training pipeline has been uploaded to GitHub.\footnote{\url{https://anonymous.4open.science/r/You_Only_Landmark_Once-20CF}}
 
We use Adam optimizer \cite{kingma2014adam} with a learning rate of $10^{-4}$, no weight decay, and default momentum parameters ($\beta_1{=}0.9$, $\beta_2{=}0.999$). To handle stagnation, we employ \texttt{ReduceLROnPlateau} as our learning rate scheduler, which has shown superior performance over fixed schedules in super-resolution and generative tasks~\cite{ledig2017photo,wang2018esrgan}.
Early stopping is based on a custom validation criterion combining perceptual quality (LPIPS~\cite{zhang2018perceptual}), structural integrity (SSIM), and signal fidelity (PSNR). SSIM and PSNR are truncated once they exceed predefined thresholds, after which the criterion becomes sensitive to LPIPS. This reflects prior findings that fidelity metrics saturate despite gains in visual realism~\cite{lim2017edsr,wang2018esrgan}, making LPIPS a more reliable signal for continued perceptual improvement.

\begin{table}[htbp]
\caption{Bilinear Downsampling Results. Quantitative comparison on aligned CelebA with 8x scaling. Our method surpasses complex prior-based architectures.}
\begin{center}
\begin{tabular}{|c|c|c|}
\hline
\textbf{Method} & \textbf{PSNR} $\uparrow$ & \textbf{SSIM} $\uparrow$ \\
\hline
Bilinear (baseline) & 20.75 & 0.574 \\
FSRGAN~\cite{chen2018fsrnet} & 22.27 & 0.601 \\
FSRNet~\cite{chen2018fsrnet} & 22.62 & 0.641 \\
VDSR~\cite{kim2016accurate} & 22.94 & 0.652 \\
ProgFSR~\cite{kim2019progressive} & 22.66 & 0.685 \\
EIPNet~\cite{lu2022edge} & 25.08 & \textbf{0.743} \\
SRFlow~\cite{lugmayr2020srflow} & 25.20 & 0.710 \\
\hline
Ours (w/o Heatmap loss) & 24.66 & 0.686 \\
Ours (w/ Heatmap loss) & \textbf{25.47} & 0.716 \\
\hline
\end{tabular}
\label{tab:metrics_30k}
\end{center}
\end{table}

\begin{table*}[htbp]
\caption{Bicubic Downsampling Results. Performance comparison on aligned CelebA ($8\times$ scaling). We categorize methods by their architectural paradigm (``Arch.'') to highlight the efficiency trade-offs. While iterative and attention-based models (e.g., MAIC, DIC) achieve high fidelity, they incur massive computational overhead. Our proposed \textit{CNN+Prior} approach offers a distinct advantage, delivering real-time efficiency with orders of magnitude fewer MACs.}
\begin{center}
\begin{tabular}{|l|c|c|c|c|c|}
\hline
\multirow{2}{*}{\textbf{Method}} & \multirow{2}{*}{\textbf{Arch.}} & \multicolumn{2}{c|}{\textbf{Complexity}} & \multicolumn{2}{c|}{\textbf{Quality} $\uparrow$} \\
\cline{3-6}
 & & \textbf{Params (M)} & \textbf{MACs (G)} & \textbf{PSNR} & \textbf{SSIM} \\
\hline
\hline
SFMNet~\cite{wang2023sfmnet} & CNN+FFT & 8.1 & 30.65 & 27.56 & 0.8074 \\
SFMNet+GAN~\cite{wang2023sfmnet} & GAN & 8.1 & 30.65 & 26.48 & 0.7662 \\
MAIC~\cite{shi2025face} & Attention & 24.2 & 19.29 & \textbf{27.68} & \textbf{0.8075} \\
DIC~\cite{ma2020deep} & Iterative & 21.8 & 45.84 & 27.37 & 0.7962 \\
\hline
\textbf{Ours (w/ Heatmap)} & \textbf{CNN+Prior} & \textbf{7.3} & \textbf{4.0} & 24.77 & 0.6924 \\ 
\textbf{Ours (w/o Heatmap)} & \textbf{CNN+Prior} & \textbf{7.3} & \textbf{4.0} & 24.01 & 0.6688 \\ 
\hline
\end{tabular}
\label{tab:metric_bicubic}
\end{center}
\end{table*}

\subsection{Effect of Heatmap Loss}
We first analyze the impact of our core contribution, namely the integration of YOLO-World-derived landmark heatmaps into a lightweight U-Net architecture for face super-resolution.
Quantitative results under bilinear downsampling are reported in Table~\ref{tab:metrics_30k}.
The baseline model, trained without heatmap-guided weighting, achieves 24.66\,dB PSNR, 0.686 SSIM and 0.9145 of MS-SSIM.
Introducing the proposed Heatmap Loss consistently improves reconstruction quality, raising performance to 25.47\,dB PSNR, 0.716 SSIM and 0.9258 MSSIM.

\begin{figure}[t] 
    \centering
    \includegraphics[width=\columnwidth]{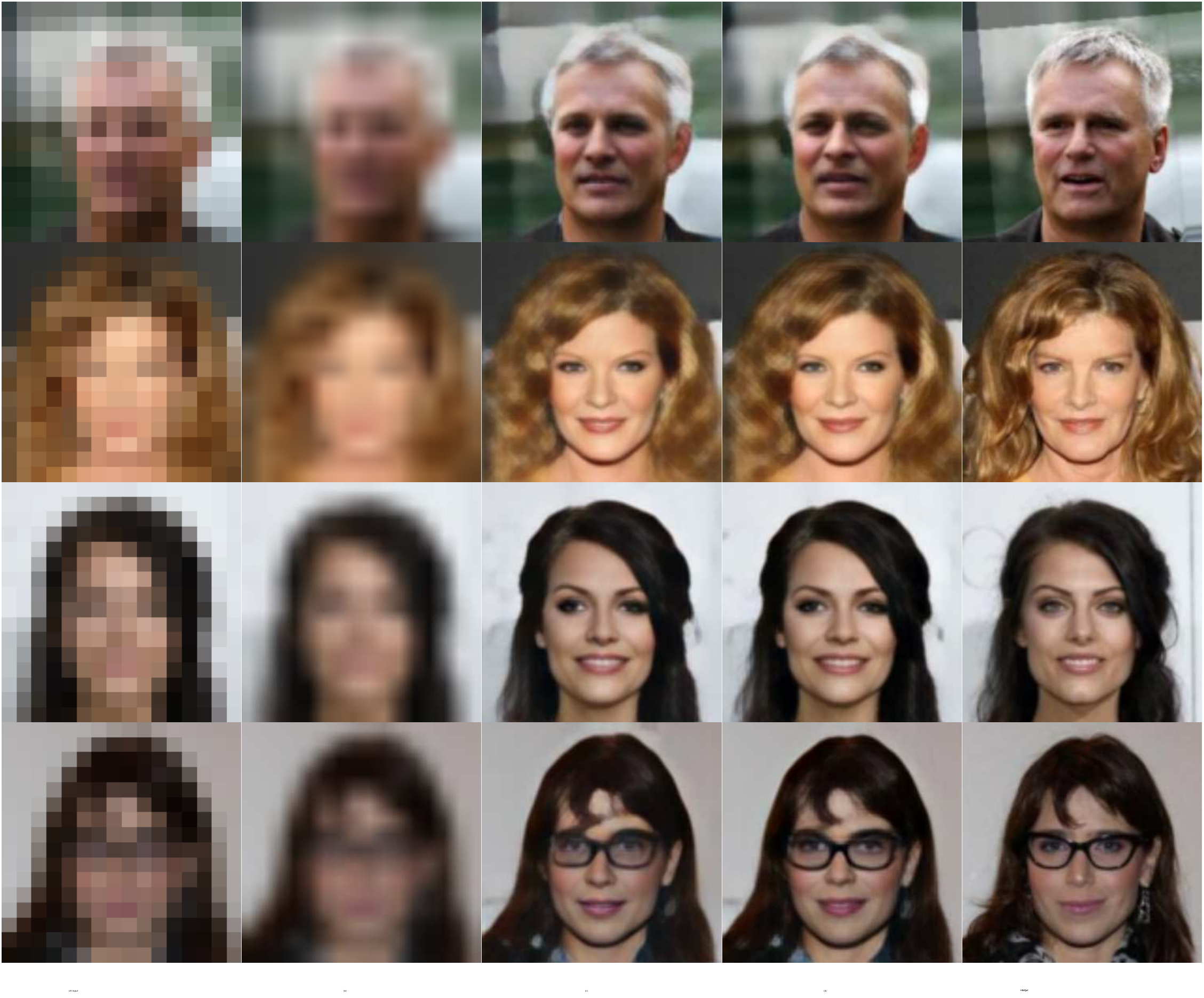}
    \caption{Qualitative comparison on aligned CelebA ($8\times$) under bilinear downsampling. In order: LR Input, (b) Bilinear interpolation, (c) Ours w/o Heatmap loss, (d) Ours w/ Heatmap loss, and HR Target. Model using heatmap loss (d) produces sharper and more coherent reconstructions in semantically important regions such as eyes and mouth, retrieving relevant details.}
    \label{fig:bilinear_faces}
\end{figure}

We attribute these improvements to the ability of detection-driven priors to emphasize semantically salient facial regions, encouraging the network to allocate capacity to perceptually critical structures. This behavior is visually supported by Fig.~\ref{fig:bilinear_faces}, where sharper facial details and reduced artifacts appear in landmark regions.
Importantly, unlike prior landmark-guided approaches based on dedicated face alignment networks\cite{kim2019progressive,bulat2018super}, our method relies on heatmaps extracted from an open-vocabulary detector.
YOLO-World provides supervision only for visible facial components, avoiding noisy guidance in the presence of occlusions and eliminating the need for task-specific retraining or auxiliary networks.

\begin{figure}[t]
    \centering
    \includegraphics[width=\columnwidth, keepaspectratio]{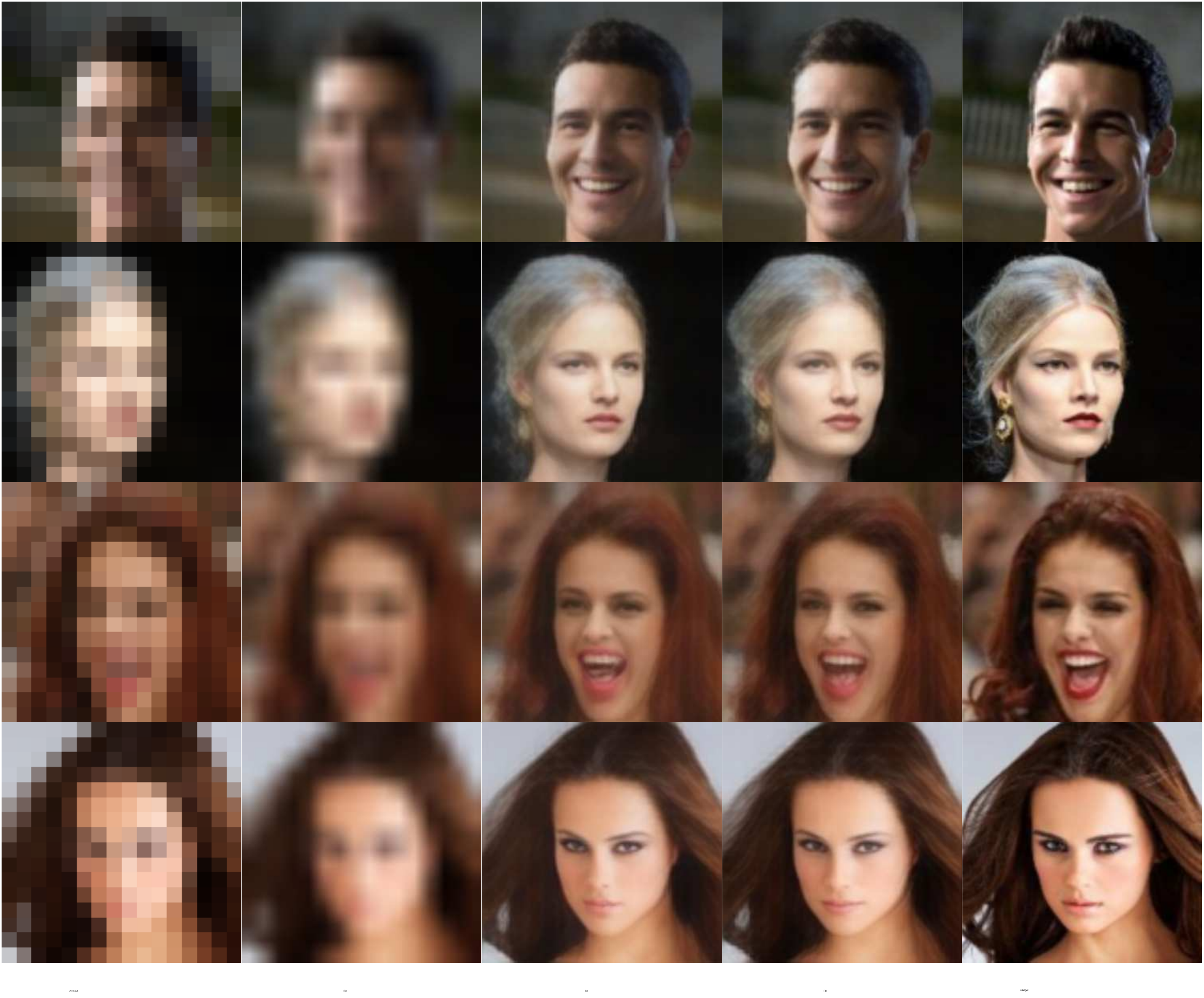}
    \vspace{-5pt}
    \caption{Qualitative comparison on aligned CelebA ($8\times$) under bicubic downsampling. In order: LR Input, (b) Bicubic interpolation, (c) Ours w/o Heatmap loss, (d) Ours w/ Heatmap loss, and HR Target. Even under more challenging settings with bicubic dowsampling, the model using heatmap loss (d) produces sharper and more coherent reconstructions, retrieving salient details in landmark regions, being more robust against generation of artifacts.}
    \label{fig:bicubic_faces}
\end{figure}

Even under the more challenging bicubic degradation setting (Table~\ref{tab:metric_bicubic}), the Heatmap Loss continues to provide consistent improvements, as shown in Fig~\ref{fig:bicubic_faces}.
These results seems to indicate that detection-driven priors from foundation models can effectively guide face super-resolution, offering a robust and lightweight alternative to specialized landmark or attention-based supervision.

\begin{table*}[t]
\caption{Efficiency analysis. Model complexity is reported in terms of trainable parameters and MACs, together with inference latency (ms) and throughput (FPS) measured on CPU and GPU. Runtime statistics are computed as the average over 200 forward passes on a fixed low-resolution input of $16 \times 16$ pixels with an $8\times$ upscaling factor. Our method exhibits substantially lower inference latency and higher throughput compared to prior face super-resolution approaches, while requiring significantly fewer MACs. *RH stands for Refinement Heads.}
\begin{center}
\begin{tabular}{|l|c|c|c|c|c|c|}
\hline
\multirow{3}{*}{\textbf{Method}} & \multicolumn{4}{c|}{\textbf{Inference Performance}} & \multicolumn{2}{c|}{\textbf{Complexity}} \\
\cline{2-7}
 & \multicolumn{2}{c|}{\textbf{CPU}} & \multicolumn{2}{c|}{\textbf{GPU}} & \textbf{Params} & \textbf{MACs} \\
\cline{2-5}
 & \textbf{Latency} $\downarrow$ & \textbf{FPS} $\uparrow$ & \textbf{Latency} $\downarrow$ & \textbf{FPS} $\uparrow$ & \textbf{(M)} $\downarrow$ & \textbf{(G)} $\downarrow$ \\
\hline
\hline
SFMNet~\cite{wang2023sfmnet} & 945.73 & 1.1 & 92.00 & 10.9 & 8.1 & 30.65 \\
MAIC~\cite{shi2025face}      & 1939.05 & 0.5 & 138.17 & 7.2 & 24.2 & 19.29 \\
DIC~\cite{ma2020deep}        & 2973.02 & 0.3 & 89.53 & 11.2 & 21.8 & 45.84 \\
\hline
\textbf{Ours}                & \underline{179.2} & \underline{5.6} & \underline{4.20} & \underline{238.4} & \underline{7.3} & \underline{4.00} \\
\textbf{Ours w/ 1 RH*}       & \textbf{79.61} & \textbf{12.6} & \textbf{2.61} & \textbf{382.4} & \textbf{7.1} & \textbf{1.28} \\
\hline
\end{tabular}
\label{tab:efficiency}
\end{center}
\end{table*}

\subsection{Efficiency Analysis}
The computational efficiency of the proposed method is analyzed by considering model complexity, runtime performance, and reconstruction quality.
Table~\ref{tab:efficiency} reports inference latency and throughput on CPU and GPU, along with the number of trainable parameters and MACs, while Table~\ref{tab:metric_bicubic} relates these efficiency metrics to reconstruction accuracy under $8\times$ bicubic downsampling on aligned CelebA.
As shown in Table~\ref{tab:efficiency}, our architecture achieves substantially lower inference latency and higher throughput compared to recent face super-resolution methods such as SFMNet, MAIC, and DIC.
In particular, the proposed model requires only 1.28--4.0 GMACs, depending on the number of refinement heads, which is an order of magnitude fewer than competing approaches.
This reduction in computational cost directly translates into real-time performance, exceeding 200 FPS on GPU and maintaining practical inference times even on CPU.

Table~\ref{tab:metric_bicubic} further highlights the trade-off between reconstruction quality and efficiency.
State-of-the-art methods based on iterative collaboration or attention mechanisms (e.g., MAIC and DIC) achieve higher PSNR and SSIM values, but at the cost of significantly increased computational complexity, with MAC counts exceeding 19G and reaching up to 45G.
In contrast, our approach attains competitive reconstruction quality (PSNR 24.77, SSIM 0.6924 and MS-SSIM 0.9189) while operating in a markedly lower computational regime, with lower latency at inference time (179ms on CPU and 4.20ms on GPU) and 4.0 GMACs.

We also run an experiment under bicubic degradation with our model equipped with the heatmap loss, decreasing the number of refinement heads from 5 to 1 to achive even higher computational efficency. We improved our efficency metrics to 79.61ms of latency and 12.6 FPS on CPU, and 2.61 ms of latency and 382.4 FPS on GPU with a total of 7.1M parameter and just 1.28 MAC at the cost slightly lower perfomances with 24.68 PSNR, 0.6878 SSIM and 0.9178 MS-SSIM.

These results indicate that, although not optimized for peak fidelity, the proposed method occupies a favorable position in the efficiency-accuracy spectrum.
By leveraging detection-driven priors extracted from a general purpose pretrained YOLO-World model and a lightweight U-Net backbone, our approach enables face super-resolution with minimal overhead, making it particularly suitable for real-time and resource-constrained scenarios.

\section{Conclusions}
This work investigated extreme face super-resolution ($8\times$, $16\times16 \rightarrow 128\times128$) through a lightweight U-Net guided by detection-driven priors. We leveraged facial landmark heatmaps extracted from the YOLO-World open-vocabulary detector to spatially weight the reconstruction loss, encouraging the model to focus on semantically important regions such as the eyes, nose, and mouth.
A key contribution is obtaining landmark information directly from ground-truth images via YOLO-World, avoiding the need to train or fine-tune auxiliary alignment networks such as FAN. Unlike prior approaches, our method eliminates inference-time heatmap generation, reducing computational overhead and simplifying the pipeline.
To isolate the effect of these priors, we employed a simple U-Net architecture, using the heatmaps solely as a weighting mask. Quantitative results show consistent improvements in PSNR, SSIM, and MS-SSIM over a baseline without priors under both bilinear and bicubic degradation, while qualitative evaluations reveal sharper, more realistic reconstructions in critical facial regions.
Overall, detection-driven priors from a general-purpose foundation model can guide extreme face super-resolution without added architectural complexity or supervision cost. With its lightweight design and low computational overhead, the framework is well suited for resource-constrained scenarios, demonstrating the potential of repurposing foundation models as zero-cost priors for efficient, image restoration under severe degradation.

\bibliographystyle{IEEEtran}
\bibliography{main}

\end{document}